\documentclass[11pt,letterpaper]{article}
\usepackage[margin=1in]{geometry}
\usepackage[T1]{fontenc}
\usepackage{amsmath,amssymb}
\usepackage{newtxtext}
\usepackage{newtxmath}
\usepackage{graphicx}
\usepackage{booktabs}
\usepackage[font=small,labelfont=bf,labelsep=period,figurename=Fig.]{caption}
\usepackage{microtype}
\usepackage{url}
\usepackage[hyperfootnotes=false,bookmarksnumbered]{hyperref}
\hypersetup{colorlinks=true, linkcolor={blue!50!black}, citecolor={blue!50!black}, urlcolor={blue!50!black},
  pdftitle={OraRefer: Segmentation-Guided Multi-Photograph Fusion for Oral Cancer Screening with Guaranteed
  Patient-Level Sensitivity},
  pdfauthor={Rupam Mukherjee, Mehakdeep Singh, Shirin Dasgupta, Subhamoy Mandal},
  pdfkeywords={Conformal prediction, multiple-instance learning, oral cancer, patient-level fusion, smartphone
  photographs}}
\graphicspath{{figures/}}

\makeatletter
\newcommand{\sg}[1]{\protected@edef\sg@v{#1}\expandafter\sg@chk\sg@v\@nil}
\def\sg@chk#1#2\@nil{\ifx$#1\else\phantom{$-$}\fi#1#2}
\makeatother
\newcommand{\hd}[1]{\begin{tabular}[b]{@{}c@{}}#1\end{tabular}}
\newcommand{\method}{OraRefer}
\newcommand{\codeurl}{https://github.com/ZenAesthete/OraRefer}
\newcommand{\blfootnote}[1]{\begingroup\renewcommand\thefootnote{}\footnotetext{#1}\endgroup}

\title{\bfseries OraRefer: Segmentation-Guided Multi-Photograph Fusion for Oral Cancer Screening with Guaranteed
Patient-Level Sensitivity}
\author{%
\begin{tabular}[t]{c}
Rupam Mukherjee$^{1\dagger}$\quad
Mehakdeep Singh$^{1\dagger}$\quad
Shirin Dasgupta$^{2}$\quad
Subhamoy Mandal$^{1}$\\[6pt]
\normalsize $^{1}$Department of Medical Science and Technology\\
\normalsize $^{2}$Dr.~B.~C. Roy Multi-Speciality Medical Research Centre\\
\normalsize Indian Institute of Technology Kharagpur, India
\end{tabular}}
\date{}

\begin{document}
\maketitle
\blfootnote{$^{\dagger}$These authors contributed equally.}
\blfootnote{This work has been submitted to the IEEE for possible publication. Copyright may be transferred without
notice, after which this version may no longer be accessible.}

\begin{abstract}
Oral cancer and the oral potentially malignant disorders (OPMD) that can precede it are visible on the oral mucosa,
which makes smartphone photographs a promising tool for screening. Deep learning models for this task classify
individual photographs. However, screening refers patients, not photographs: a patient is usually photographed
several times, and the referral decision must miss no more than a set share of patients with disease. We propose
\method, a patient-level screening model that reads each photograph with a segmentation-guided foundation model
that conditions its class decision on the predicted lesion, combines all photographs of a patient with risk factors, and refers through a threshold with a finite-sample
sensitivity guarantee. The model localizes the mouth itself, so inference needs no annotation. On
3000 photographs of 714 patients, patient-level fusion reaches an AUROC of
0.919 and outperforms attention-based multiple-instance learning. At a 90\%
sensitivity target, \method\ detects 95\% of patients with OPMD or cancer and misses
3 of 61 patients with cancer, at a specificity of 0.610. Because the threshold is calibrated without
retraining, each clinic can set its own guaranteed operating point on a local calibration sample.

\end{abstract}

\section{Introduction}
\label{sec:intro}

Cancers of the lip and oral cavity caused 389485 new cases and 188230 deaths worldwide in
2022~\cite{bray2024globocan}. They can be preceded by oral potentially malignant disorders (OPMD), visible mucosal
abnormalities with an increased risk of cancer~\cite{warnakulasuriya2021opmd}. A photograph of the mouth can
therefore show the disease that screening must find, and smartphones bring such photographs within reach of primary
care. The screening decision, however, concerns a patient, not a photograph. A patient is often photographed
several times, carries risk factors such as tobacco and betel quid use~\cite{secretan2009carcinogens}, and must be
referred if any lesion is suspicious~\cite{nice2015ng12}. This calls for a patient-level model whose sensitivity holds on the next patients it sees,
not only on average.

With the release of public datasets, deep learning on oral photographs has advanced. The dataset of Piyarathne et
al.~\cite{piyarathne2024oral} provides 3000 smartphone photographs of 714
patients with lesion polygons and risk factors. On these data, classifiers have used convolutional and transformer
backbones~\cite{ormeno2026imbalance}, oral-cavity segmentation followed by image--metadata
fusion~\cite{devindi2024multimodal}, embedding-level fusion with clinical data~\cite{ormeno2026multimodal}, and
stacked ensembles with per-patient posterior smoothing~\cite{hazra2026uncertainty}. Two of these studies~\cite{ormeno2026imbalance,ormeno2026multimodal}
split the data by image, which places photographs of one patient in both the training and the test set. More importantly, none
of them turns all photographs of a patient into one referral decision with controlled sensitivity. General tools exist
for each half of the problem: attention-based multiple-instance learning pools instances into a bag
label~\cite{ilse2018attention}, including patients in oral cytology~\cite{mukherjee2026raamil}, and conformal and Neyman--Pearson thresholds bound the errors of a fixed
classifier~\cite{tong2018neyman,angelopoulos2023gentle}, with conformal triage applied to medical
imaging~\cite{angelopoulos2024triage}. A patient-level bound, however, requires out-of-sample patient scores at both
stages of the model.

\begin{figure}[tbp]
\centering
\includegraphics{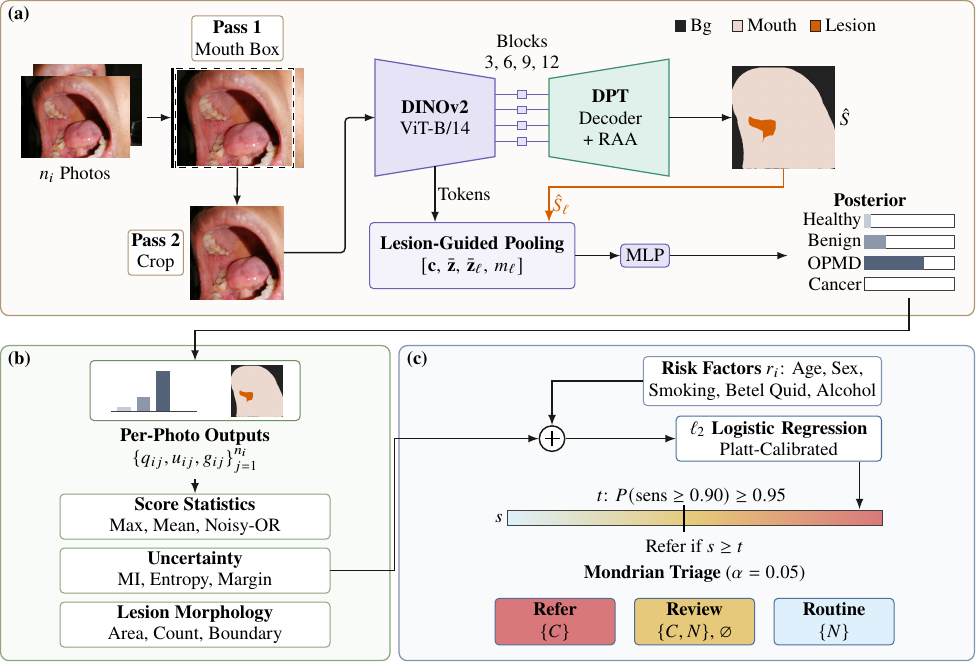}
\caption{\method\ needs no annotation at inference, fuses a patient's photographs with risk factors, and refers with guaranteed sensitivity. (a)~Pass 1 predicts the mouth box (dashed); pass 2 segments the crop ($\hat{S}$) and pools its tokens with the lesion map $\hat{S}_{\ell}$. Shown: a held-out patient with OPMD~\cite{piyarathne2024data}. (b)~Per-photograph outputs are aggregated per patient. (c)~The threshold $t$ on the patient score $s$ and Mondrian triage ($C$: OPMD or cancer, $N$: neither) are calibrated on held-out patients without refitting.}
\label{fig:pipeline}
\end{figure}

We propose \method, a screening model that goes from a patient's raw photographs to a referral with
guaranteed sensitivity (Fig.~\ref{fig:pipeline}). First, we introduce a
segmentation-guided image model: one network, trained jointly to segment the mouth and the lesion and to classify the
photograph, that conditions its class decision on the predicted lesion. Where earlier work segments the oral cavity in
a separate preprocessing network~\cite{devindi2024multimodal}, our segmentation training shapes the features the
classifier reads, and the same network crops the mouth from the full frame, so no annotation is needed at test time. Second, we introduce a patient-level referral
layer that fuses all photographs of a patient with risk factors and refers through a distribution-free sensitivity
threshold and Mondrian conformal triage, both calibrated on held-out patients without retraining. Third, we show on
714 patients that \method\ outperforms frozen foundation-model probes, untrained aggregation, and attention-based multiple-instance learning, and detects 95\% of patients with OPMD or cancer,
including 58 of 61 with cancer, against a 90\% target.
{}
{}
\section{Method}
\label{sec:method}

\method\ has three parts (Fig.~\ref{fig:pipeline}). An image model reads each photograph in two passes and returns
class probabilities, a lesion mask, and an uncertainty estimate. A fusion model turns the outputs for all photographs
of a patient, together with the patient's risk factors, into one score. A threshold calibrated on held-out patients
then turns the score into a referral. We call a patient referable if any of their photographs shows OPMD or cancer.

\subsection{Segmentation-Guided Image Model}
\label{sec:image}
The encoder is \mbox{DINOv2} \mbox{ViT-B/14}~\cite{oquab2024dinov2}, fine-tuned end to end with layer-wise
learning-rate decay. A DPT decoder~\cite{ranftl2021dpt} reassembles four encoder blocks into a dense map that segments
background, mouth, and lesion. Within the decoder, region-affinity attention~\cite{mukherjee2026raamil} aggregates each reassembled token over its
$3\times3$ neighborhood with learned affinity weights. The decoder guides the classifier: the head receives the class token $\mathbf{c}$, the
mean token $\bar{\mathbf{z}}$, the lesion-weighted token mean $\bar{\mathbf{z}}_{\ell}$, and the predicted lesion mass
$m_{\ell}$. The model is trained with
\begin{equation}
\mathcal{L} = \mathcal{L}_{\mathrm{seg}} + \lambda_{\mathrm{cls}}\mathcal{L}_{\mathrm{cls}}
+ \lambda_{\mathrm{con}}\mathcal{L}_{\mathrm{con}} + \lambda_{\mathrm{raa}}\mathcal{L}_{\mathrm{raa}}
+ \mathcal{L}_{\mathrm{aux}},
\label{eq:loss}
\end{equation}
where $\mathcal{L}_{\mathrm{seg}}$ combines cross-entropy, Tversky ($\alpha_{\mathrm{T}}=0.3$,
$\beta_{\mathrm{T}}=0.7$), and boundary losses, $\mathcal{L}_{\mathrm{cls}}$ is label-smoothed cross-entropy, $\mathcal{L}_{\mathrm{con}}$ is supervised contrastive learning~\cite{khosla2020supcon} on projected pooled features, and $\mathcal{L}_{\mathrm{raa}}$ penalizes affinity across annotated lesion
boundaries ($\lambda_{\mathrm{cls}}=0.5$, $\lambda_{\mathrm{con}}=0.1$,
$\lambda_{\mathrm{raa}}=0.05$). $\mathcal{L}_{\mathrm{aux}}$ holds decaying auxiliary losses on an
intermediate decoder output and a second classifier head, and a term that ties the healthy-class probability to the
predicted lesion area.

\subsection{Self-Localization at Test Time}
\label{sec:selfloc}
Training mixes full frames with mouth crops, so the model also reads uncropped photographs. At test time, a first pass
on the full frame predicts the mouth mask, whose bounding box, padded by 8\%, is the crop for the
second pass. Inference therefore needs no annotation.

\subsection{Patient-Level Fusion}
\label{sec:fusion}
Each photograph yields the four-class posterior, the lesion area, the number of lesion components and their boundary
complexity, and the mutual information of 30 Monte Carlo dropout samples, plus the entropy and top-two margin of the posterior. Training samples the classes evenly, with cancer twice as often, so for fusion we
correct the posterior from these sampling frequencies to the training class frequencies~\cite{saerens2002adjusting} and temperature-scale it on
internal validation patients~\cite{guo2017calibration}. For patient $i$ with photographs $j=1,\dots,n_i$, let $q_{ij}$
be the corrected probability of OPMD or cancer. We summarize the photographs by the maximum, mean, and top-two mean of
$q_{ij}$, the fraction with $q_{ij}>0.5$, and the noisy-OR $1-\prod_j(1-q_{ij})$. We add the maximum and mean
cancer probability and lesion area, the lesion morphology $g_{ij}$ and the uncertainty $u_{ij}$, and the risk factors $r_i$ (age, sex,
smoking, betel quid, and alcohol use). An $\ell_2$-regularized logistic regression with inner cross-validated penalty and Platt
calibration maps these features to the patient score $s_i$.

The fusion model must see out-of-sample image outputs in training, as at test time, so we cross-fit the image model
at two levels. With $K$ patient folds, the model $M_{-\{f,f'\}}$ is trained without folds $f$ and $f'$.
For held-out fold $f$, fusion training features of fold $f'\neq f$ come from $M_{-\{f,f'\}}$, and test patients in fold $f$ are scored through each of the $K-1$ models $M_{-\{f,f'\}}$ and averaged in logit space. This way, no image or
fusion model sees fold $f$ before it is scored.

\subsection{Referral with Guaranteed Sensitivity}
\label{sec:referral}
Let $s_{(1)}\le\dots\le s_{(n)}$ be the scores of $n$ referable calibration patients, and let $\tau$ be the target
sensitivity. For exchangeable patients with continuous scores, the miss probability of the threshold $s_{(k)}$, the
probability that a new referable patient scores below it, follows $\mathrm{Beta}(k,n-k+1)$ over calibration sets. The guaranteed-sensitivity threshold is $t=s_{(k^\ast)}$ with
\begin{equation}
k^\ast=\max\{k:\Pr[\mathrm{Bin}(n,1-\tau)\ge k]\ge 1-\delta\},
\label{eq:threshold}
\end{equation}
so that \mbox{$\Pr[\mathrm{sensitivity}\ge\tau]\ge 1-\delta$} for new patients, an order-statistic bound as in
Neyman--Pearson classification~\cite{tong2018neyman}. This threshold exists for $n\ge\lceil\ln\delta/\ln\tau\rceil$, that is
$n\ge29$ for $\tau=0.90$ and $\delta=0.05$. The conformal threshold $t=s_{(\lfloor(n+1)(1-\tau)
\rfloor)}$ instead guarantees sensitivity $\ge\tau$ in expectation~\cite{angelopoulos2023gentle} and refers fewer
patients.

Mondrian conformal triage~\cite{vovk2012conditional} adds a third option. For each class, referable or not, a new
patient's p-value ranks its non-conformity, $1-s$ for referable and $s$ for not, among the calibration patients of
that class, and the prediction set keeps every class with p-value above $\alpha$. A set with only the referable class means referral; with only the other class, routine care; with both or neither, clinician review. Coverage per class is at least $1-\alpha$ in expectation, so on average at most a share $\alpha$ of
referable patients is sent to routine care.

Both thresholds are calibrated inside the held-out fold. Each fold is split into five label-stratified parts, and each
part is scored with the threshold calibrated on the other four. Calibration thus spends no training patient and needs
no further model fit.
{}
\section{Experiments}
\label{sec:experiments}

\subsection{Data and Protocol}
We use the dataset of Piyarathne et al.~\cite{piyarathne2024oral}: 3000 photographs of
714 patients, taken with mobile phones by dental surgeons at a teaching hospital in Sri Lanka.
Under the supervision of oral medicine specialists, each photograph is labeled healthy (729),
benign (748), OPMD (1394), or cancer (129) and annotated
with mouth and lesion polygons. Histopathology confirmed all cancer and 56.4\% of OPMD diagnoses~\cite{piyarathne2024oral}. Each patient has a median of four photographs
(range 1--18) and records of age, sex, and three binary risk factors.
381 patients are referable, 61 of them with cancer, and 333 are not.

All results come from patient-level five-fold cross-validation stratified by patient category and risk factors.
Two patient pairs share an identical photograph; each pair stays in one fold and one calibration part. All model selection happens inside the training
folds. The image model uses one configuration for all folds; inside each training set, 17\% of
patients are held out to select the checkpoint, the temperature, and the lesion-mask threshold, and the fusion penalty
is chosen by inner cross-validation. The image model trains for 20 epochs at 448\,px with batch size 16, AdamW, and base learning
rate \ensuremath{3\times10^{-4}}, on a 16\,GB V100. Thresholds target $\tau=0.90$ with $\delta=0.05$, and triage uses
$\alpha=0.05$. The 25 calibration sets hold 60--62
referable patients each, 8--12 of them with cancer, and triage shares are averaged over 20 random calibration re-splits of
each fold.

\subsection{Comparisons and Metrics}
At the image level, removing segmentation guidance yields a \mbox{DINOv2}
classifier fine-tuned on the same predicted crops with the same schedule, without segmentation losses, lesion-guided pooling, or region-affinity attention. We also remove
lesion-guided pooling, region-affinity attention, or supervised contrastive learning alone, give the second pass the
annotated mouth box, and train linear probes on frozen full-photograph \mbox{DINOv2-B}, \mbox{DINOv2-L}, and \mbox{SigLIP2}~\cite{tschannen2025siglip2}.

Patient-level baselines form three groups: logistic regression on image features or risk factors alone; untrained aggregation by the maximum, mean, and noisy-OR of $q_{ij}$; and trained learners: gradient boosting, late-fusion networks with and without gating by image uncertainty, and ABMIL~\cite{ilse2018attention} and
CAR-MIL~\cite{chraki2026carmil}. The attention models receive the same per-photograph outputs and risk factors plus
the first 32 principal components of frozen \mbox{DINOv2-B} embeddings of the full photograph, and CAR-MIL adds a counterfactual attention regularizer
to ABMIL. Like \method, every trained baseline is Platt-calibrated, and the neural ones stop early on internal
validation patients.

We report four-class macro AUROC and macro-F1 of the classifier posterior, lesion Dice on photographs with a lesion,
and predicted mouth-box IoU, all from out-of-fold predictions of the five-fold image models. Patient-level results give AUROC and the
sensitivity, specificity, and referral rate of the guaranteed-sensitivity threshold. Brackets are 95\% intervals:
2000 patient-cluster bootstrap resamples for image metrics, DeLong~\cite{delong1988comparing} for patient AUROC, and a bootstrap recalibrating the thresholds in each resample for operating points.
{}
\section{Results}
\label{sec:results}

\begin{table}[tbp]
\centering
\caption{Segmentation guidance drives the image model, and self-localization matches annotated crops. $\Delta$: change
in macro AUROC from \method; folds lower: number of the five folds in which it falls. A second training seed changes
macro AUROC by $-$0.005.}
\label{tab:image}
\small
\begin{tabular}{@{}lclcccc@{}}
\toprule
 & \hd{Macro\\AUROC} & \multicolumn{1}{c}{\hd{$\Delta$\\{}[95\% interval]}} & \hd{Folds\\lower} & \hd{Cancer\\AUROC}
 & \hd{Macro-F1} & \hd{Lesion\\Dice} \\
\midrule
\method & 0.893 & \multicolumn{1}{c}{--} & -- & 0.942 & 0.704 & 0.580 \\
\quad w/o segmentation guidance & 0.869 & \sg{$-$0.023}~[\sg{$-$0.033},~\sg{$-$0.015}] & 5 & 0.923 & 0.688 & -- \\
\quad w/o lesion-guided pooling & 0.884 & \sg{$-$0.009}~[\sg{$-$0.016},~\sg{$-$0.002}] & 4 & 0.919 & 0.703 & 0.587 \\
\quad w/o region-affinity attention & 0.893 & \sg{0.001}~[\sg{$-$0.005},~\sg{0.007}] & 1 & 0.943 & 0.703 & 0.561 \\
\quad w/o supervised contrastive loss & 0.885 & \sg{$-$0.007}~[\sg{$-$0.014},~\sg{$-$0.002}] & 4 & 0.934 & 0.705 & 0.581 \\
\quad with annotated mouth box & 0.891 & \sg{$-$0.001}~[\sg{$-$0.003},~\sg{0.000}] & 3 & 0.940 & 0.707 & 0.583 \\
\midrule
Linear probe, frozen \mbox{DINOv2-B} & 0.809 & \sg{$-$0.084} & & & & \\
Linear probe, frozen \mbox{DINOv2-L} & 0.823 & \sg{$-$0.069} & & & & \\
Linear probe, frozen \mbox{SigLIP2} & 0.817 & \sg{$-$0.076} & & & & \\
\bottomrule
\end{tabular}
\end{table}
\begin{table}[t]
\centering
\caption{\method\ has the highest patient-level AUROC. Guaranteed-sensitivity operating points ($\tau=0.90$, $\delta=0.05$); Ref.: referral rate. ABMIL and CAR-MIL AUROCs differ only in the fourth decimal.}
\label{tab:patient}
\small
\begin{tabular}{@{}lcccc@{}}
\toprule
 & AUROC & Sens. & Spec. & Ref. \\
\midrule
Risk factors only & 0.760~[0.724,~0.795] & 0.958 & 0.201 & 88\% \\
Image only & 0.909~[0.888,~0.930] & 0.958 & 0.562 & 72\% \\
Max of $q_{ij}$ & 0.878~[0.853,~0.903] & 0.950 & 0.538 & 72\% \\
Mean of $q_{ij}$ & 0.889~[0.865,~0.914] & 0.958 & 0.502 & 74\% \\
Noisy-OR & 0.851~[0.824,~0.879] & 0.961 & 0.387 & 80\% \\
ABMIL & 0.890~[0.867,~0.914] & 0.953 & 0.508 & 74\% \\
CAR-MIL & 0.890~[0.867,~0.914] & 0.953 & 0.517 & 73\% \\
Gradient boosting & 0.916~[0.895,~0.936] & 0.953 & 0.556 & 72\% \\
Late fusion & 0.915~[0.895,~0.935] & 0.942 & 0.592 & 69\% \\
\quad + gating & 0.909~[0.887,~0.930] & 0.950 & 0.574 & 71\% \\
\method & \textbf{0.919~[0.899,~0.939]} & 0.950 & 0.610 & 69\% \\
\bottomrule
\end{tabular}
\end{table}

\subsection{Image Model}
Segmentation guidance drives the image model (Table~\ref{tab:image}). \method\ reaches a four-class macro AUROC of
0.893 and a macro-F1 of 0.704. This exceeds the best published four-class result under
patient-grouped evaluation, a macro-F1 of 0.662 on a curated subset of 2438 photographs cropped to the
lesion~\cite{hazra2026uncertainty}, whereas \method\ reads all 3000 photographs uncropped. \method\
also outperforms the frozen foundation-model probes by
0.069--0.084 in macro AUROC. Without segmentation guidance, macro AUROC falls by
0.023 on the same crops. The drop is consistent across all five folds, at five times the seed-to-seed change. Lesion-guided pooling accounts for 0.009 of this drop and the segmentation
training for the rest, and supervised contrastive learning adds a further 0.007; pooling and
contrastive learning each help in four of five folds. Region-affinity attention raises lesion Dice from
0.561 to 0.580. The gain from segmentation guidance carries through to the referral decision. For each ablated
image model, we refit the fusion model on its out-of-fold outputs, a single level of cross-fitting under which the full
model reaches 0.913 instead of 0.919. Removing segmentation guidance lowers patient AUROC
to 0.894 (DeLong $p=0.005$),
four times the patient-level seed-to-seed change (0.005), whereas no other component changes it by more than 0.008.

Self-localization costs no classification accuracy. The predicted mouth box has a median IoU of 0.981
with the annotated box, and an IoU of at least 0.5 on 2998 of 3000 photographs. Replacing
it with the annotated box changes macro AUROC by only $-$0.001.

\subsection{Patient-Level Fusion}
Learned fusion outperforms untrained pooling of the photograph scores (Table~\ref{tab:patient}). Even without risk
factors, the image-only model reaches an AUROC of 0.909, above the untrained aggregates
(DeLong $p\le0.005$). Risk factors add information beyond the photographs, which a likelihood-ratio
test of adding them to a logistic model of the image-only score confirms ($\chi^2(5)=34.4$, $p<0.001$), and with them \method\ reaches
0.919, the highest AUROC in Table~\ref{tab:patient}, above each untrained aggregate (DeLong
$p<0.001$). It outperforms ABMIL and CAR-MIL by 0.029 ($p=0.001$ each,
Holm-adjusted over 13 comparisons with the other trained models in Table~\ref{tab:patient} and
six variants of \method), although both attention models also receive frozen \mbox{DINOv2} embeddings, and at the
guaranteed sensitivity it reaches a specificity of 0.610, compared with 0.562 for image
evidence alone and 0.508 for ABMIL. More flexible learners do not improve on the calibrated logistic
regression: gradient boosting reaches 0.916, and the late-fusion network 0.915, or
0.909 when gated by image uncertainty. Neither photograph counts nor the four-model test scoring drive
these results: count-free summaries (means and fractions) give 0.912, and scoring each test patient,
like each training patient, with a single image model (trained on the four training folds) gives 0.917, both above every untrained aggregate and attention model.

\begin{figure}[tbp]
\centering
\includegraphics{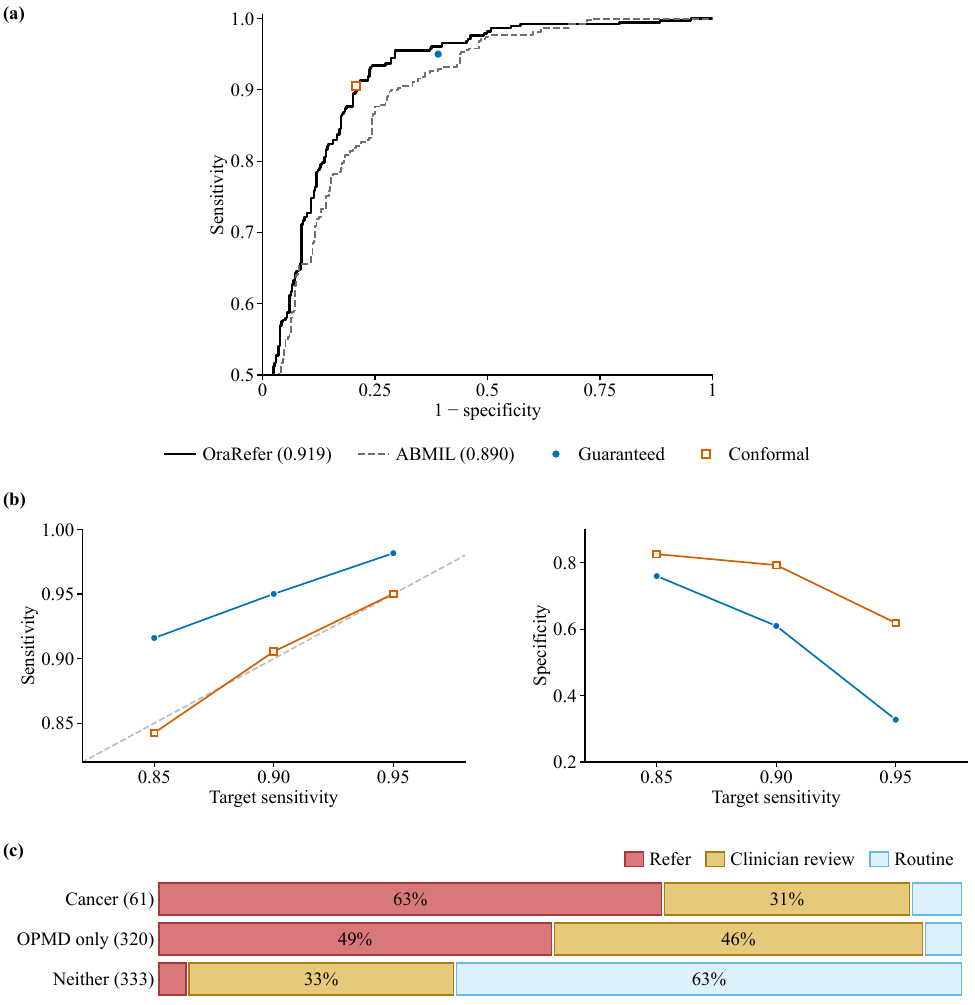}
\caption{The guaranteed-sensitivity threshold exceeds every target, and Mondrian triage sends few referable patients
to routine care. (a)~Patient-level ROC curves (AUROC) with the operating points of both thresholds at $\tau=0.90$.
(b)~Sensitivity (left; dashed: sensitivity $=\tau$) and specificity (right) of both thresholds against the target
$\tau$. (c)~Triage ($\alpha=0.05$) of patients grouped by their most
severe photograph label, mean over 20 calibration re-splits.}
\label{fig:referral}
\end{figure}

\subsection{Referral}
The guaranteed-sensitivity threshold detects 0.950~[0.934,~0.966] of referable patients against a target of 0.90, at
a specificity of 0.610~[0.470,~0.710]. It exceeds every target (Fig.~\ref{fig:referral}(b)), by the margin that a guarantee at
$1-\delta=0.95$ requires with 60--62 referable calibration patients.
Patients with cancer are detected at the same rate as all referable patients, for whom the threshold is calibrated:
58 of 61, a sensitivity of 0.951~[0.866,~0.986], and 304 of 320 with OPMD only. The conformal threshold, by contrast, lands on its
target (0.906 at 0.90) and trades sensitivity for a higher specificity of 0.793 (Fig.~\ref{fig:referral}(b)). Mondrian
triage refers 29\% of patients directly, sends 39\%
to clinician review, and returns 32\% to routine care, which receives 5.1\%
of referable patients (4.3\%--6.3\% over re-splits), consistent with $\alpha=0.05$ (Fig.~\ref{fig:referral}(c)).

The missed cancers are classification errors, not localization errors. All three were localized and segmented well,
with a best-photograph lesion Dice of at least 0.825, yet the classifier gave none of their photographs a
$q_{ij}$ above 0.380. Missed patients with OPMD only, in contrast, show small predicted lesions: their largest covers a median 0.04 of the predicted mouth, against
0.32 for detected ones.
{}
\section{Conclusion}
\label{sec:conclusion}

\method\ turns a patient's raw smartphone photographs into a referral with guaranteed sensitivity. Segmentation
guidance drives its image model, and self-localization removes the need for test-time annotation. Patient-level
fusion with risk factors outperforms untrained aggregation and attention-based multiple-instance learning. Since
the thresholds are calibrated without retraining, a new clinic can set its own guaranteed
operating point from a representative local sample of patients with OPMD or cancer. These retrospective results motivate prospective evaluation in community screening.
{}

\section*{Code Availability}
The code of \method\ is available at \url{\codeurl}.

\section*{Compliance with Ethical Standards}
This study was conducted using human subject data, available on request~\cite{piyarathne2024data}. Ethical approval
was not required.

\section*{Acknowledgments}
The authors have no relevant financial or non-financial interests to disclose.

\bibliographystyle{IEEEbib-etal}
\bibliography{refs}

\end{document}